\documentclass{article}

\usepackage{arxiv}

\usepackage[utf8]{inputenc} % allow utf-8 input
\usepackage[T1]{fontenc}    % use 8-bit T1 fonts
\usepackage{hyperref}       % hyperlinks
\usepackage{url}            % simple URL typesetting
\usepackage{booktabs}       % professional-quality tables
\usepackage{amsfonts}       % blackboard math symbols
\usepackage{nicefrac}       % compact symbols for 1/2, etc.
\usepackage{microtype}      % microtypography
\usepackage{lipsum}
\usepackage{graphicx}
\graphicspath{ {./images/} }
\usepackage{graphicx}
\usepackage[symbol]{footmisc}
\usepackage{subfig}
\usepackage[ruled,vlined]{algorithm2e}
\usepackage{algorithmic}
\usepackage{stackengine}
\usepackage{mathtools}

\usepackage{amsfonts}
\usepackage[export]{adjustbox}
\usepackage[numbers]{natbib}
\usepackage{tabularx}
\usepackage{multirow}
\usepackage{tabularx}
\usepackage{soul}
\usepackage{xcolor}

\title{DropKAN: Regularizing KANs by masking post-activations}

\author {
    Mohammed Ghaith Altarabichi\\
    Center for Applied Intelligent Systems Research\\
    Halmstad University, Sweden\\
    mohammed\_ghaith.altarabichi@hh.se
}

\begin{document}
\maketitle
\begin{abstract}
%Kolmogorov-Arnold Networks (KANs) are recently proposed as an alternative to Multi-Layer Perceptrons (MLPs). KANs rely on learnable activations represented as splines on edges, while KANs' node simply sum the incoming signals. We show in this work that Dropout a popular method to prevent overfitting in MLPs behaves differently during forward and backward passes when applied to KANs. We propose accordingly DropKAN a variant of Dropout designed to regularize KANs by randomly dropping the output of some activations during feedforwad stage instead of dropping nodes.  
We propose DropKAN (Dropout Kolmogorov-Arnold Networks) a regularization method that prevents co-adaptation of activation function weights in Kolmogorov-Arnold Networks (KANs). DropKAN functions by embedding the drop mask directly within the KAN layer, randomly masking the outputs of some activations within the KANs' computation graph. We show that this simple procedure that require minimal coding effort has a regularizing effect and consistently lead to better generalization of KANs.

We analyze the adaptation of the standard Dropout with KANs and demonstrate that Dropout applied to KANs' neurons can lead to unpredictable behavior in the feedforward pass. We carry an empirical study with real world Machine Learning datasets to validate our findings. Our results suggest that DropKAN is consistently a better alternative to using standard Dropout with KANs, and improves the generalization performance of KANs. Our implementation of DropKAN is available at: \url{https://github.com/Ghaith81/dropkan}. 
\end{abstract}

% keywords can be removed
%\keywords{First keyword \and Second keyword \and More}

\section{Introduction}
Kolmogorov-Arnold Networks (KANs) \cite{liu2024kan} are recently proposed as an alternative to Multi-Layer Perceptrons (MLPs). The computation graph of Kolmogorov-Arnold Networks (KANs) is different form the standrad Multi-Layer Perceptrons (MLPs) in two fundamental ways: 1) On \textit{edges}: KANs use trainable activation functions, unlike MLPs that rely on linear weights. 2) On \textit{neurons} ("nodes"): KANs sum the incoming signals, different to MLPs that apply non-linear activation functions e.g., ReLU. These changes to the computation graph indicates that many of the techniques used in MLPs might not directly transfer to KANs, or at least may not necessarily give the same desired effect. In this work we explore whether KANs could benefit from using Dropout \cite{srivastava2014dropout} for regularization, and propose DropKAN based on our analysis as a method to regularize KANs by randomly masking the outputs of activations. DropKAN is efficient at regularizing KANs and is easy to incorporate within any implementation of KANs.

Our contributions in this paper can be categorized in two folds. First, we analyze the behavior of Dropout applied to KANs and show that it behaves differently from how it was originally designed to operate with MLPs. Our second and main contribution is proposing DropKAN, an alternative to Dropout applied to KANs, as we show that DropKAN consistently outperforms Dropout with KANs using a number of real world datasets. 
%The remainder of the article is organized as follows. In Section 2, we discuss related work on studying and \hl{tuning} randomness techniques in DNNs. In Section 3, we explain our \hl{hyperparameter optimziation approach} and \hl{define the two proposed new randomization techniques:} loss function noise and gradient dropout. The experiments are presented in Section 4, and the results are further discussed in Section 5. Conclusions are listed in Section 6, while limitations and future work are covered in Section 7. 
\section{Motivation}
\begin{figure*}[t]
\centering
\subfloat{\includegraphics[width=0.3\textwidth]{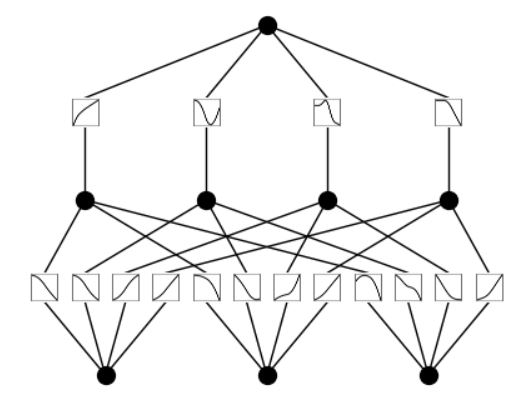}}
\subfloat{\includegraphics[width=0.3\textwidth]{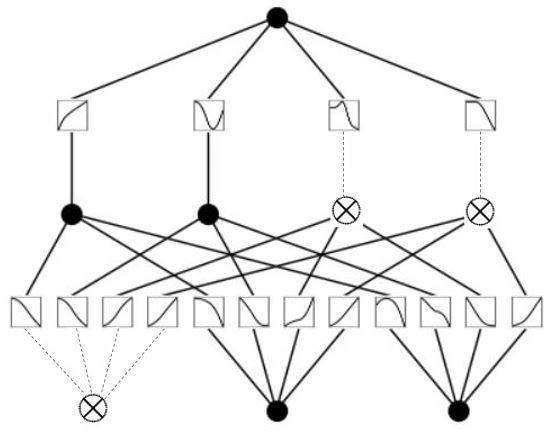}}
\subfloat{\includegraphics[width=0.3\textwidth]{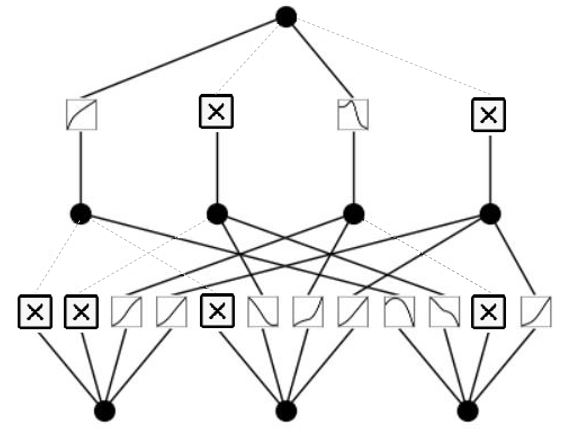}}
\caption{A standard KAN network (on the left), a KAN network regularized using Dropout applied between the KAN layers by masking neurons/nodes, with dashed gray lines indicating zero inputs to the connected activations  (middle), a KAN network regularized using DropKAN by masking post-activations, where dashed gray lines indicate zero inputs to the connected neurons (right).}
\label{fig1}
\end{figure*}

We start by formalizing the definition of a KAN layer and the adaptation of Dropout to KANs. We denote the input dimension of the $l^{th}$ layer of a KAN model consisting of $L$ layers by $n_{l}$. The activation functions connecting the layer $l$ to the following layer $l+1$ can be expressed as a 1D matrix of functions:
\begin{equation}    
\begin{aligned}
\Phi_{l} = \{\phi_{l,j,i}\},  \quad
      l = 0, 2, \ldots, L-1, 
      i = 1, 2, \ldots, n_{l}, 
      j = 1, 2, \ldots, n_{l+1}
\end{aligned}
\end{equation}
where $\phi_{l,j,i}$ is an arbitrary function represented as a spline with trainable parameters. We define $x_{l,i}$ as the pre-activation (input) of the function $\phi_{l,j,i}$; the post-activation of $\phi_{l,j,i}$ is denoted by $\tilde{x}_{l,j,i} = \phi_{l,j,i}(x_{l,i})$. The neurons in the KAN layer performs a sum of all incoming post-activations to output $\tilde{x}_{l,j,i}$.
\begin{equation}
\begin{aligned}
x_{l+1,j} = \sum_{i=1}^{n_{in}} \tilde{x}_{l,j,i} =  \sum_{i=1}^{n_{in}} \phi_{l,j,i}(x_{l,i}), 
j = 1, 2, \ldots, n_{l+1}
\label{kan}
\end{aligned}
\end{equation}
By adding a Dropout layer between the KAN layers $l$ and $l+1$ as in Figure~\ref{fig1}, a binary dropout mask \(m_{j}\) is applied to the $l$ layer outputs' $x_{l+1,j}$ in training time. The mask is usually sampled independently from a Bernoulli distribution with a probability \(p\) of being 1 (indicating that the neuron $j$ is retained) and \(1-p\) of being 0 (indicating that the neuron $j$ is dropped). The output of the KAN layer with the inclusion of Dropout becomes:
\begin{equation}
x^{\prime}_{l+1,j} = \frac{m_{j}x_{l+1,j}}{1-p}, \quad j = 1, 2, \ldots, n_{l+1}
\label{dropout}
\end{equation}
The output of the Dropout layer in Equation~\ref{dropout} is usually scaled-up by a factor of $\frac{1}{1-p}$, we denote this factor by $s$. This procedure is done to compensate the effect of dropping out some nodes, with the rational of ensuring that a similar level of signal will continue to propagate through the network with the presence of Dropout in training time. %However, we explain next that this procedure is unnecessary with KANs. In our adaptation of Dropout with KANs we use the following instead of Equations~\ref{dropout}:
%\begin{equation}
%x_{l+1,j} = m_{j}x_{l+1,j}, \quad j = 1, 2, \ldots, n_{l+1}
%\label{mydropout}
%\end{equation}

\subsection{Why is Dropout problematic with KANs?}
The key motivation of Dropout is to prevent co-adaptations \cite{hinton2012improving} of weights by sampling a thinned network through dropping out nodes. If a node in a MLP is dropped by a Dropout layer then all the weights connected to it will take no part in the feedforward and backward passes during the training step. The weights of the dropped neuron in a MLP have no impact on feedforward because they are multiplied by an input of zero from the dropped node, while on the backward pass they have no influence on the loss calculation, and will consequently get a gradient of zero in that training step.

However, applying Dropout to the outputs of a KAN layer is not enough to exclude the masked nodes from actively participating in the feedforward and backward passes. The zero outputs from the masked nodes will be used as inputs to the activation functions $\Phi_{l+1}$ in the following layer $l+1$, given that for an arbitrary activation function $\phi^{\prime}$ from $\Phi_{l+1}$, the output of $\phi^{\prime}(x=0)$ is not necessarily zero, the nodes will still propagate the corresponding value of $\phi^{\prime}(x=0)$ into the network during feedforward. Consequently, the weights of activation function (the spline coefficients) $\phi^{\prime}$ in the layer $l+1$ following the dropout will also be updated in the backward pass.

Another key issue with applying Dropout to KANs is the scaling-up procedure. As we have observed in Equation~\ref{dropout}, the outputs of the kept nodes are scaled by a factor of $s$. This procedure is ineffective with KANs as the arbitrary activation function from $\Phi_{l+1}$ are not necessarily homogeneous function of degree 1, $\phi^{\prime}(sx) \neq s\phi^{\prime}(x)$. The behavior of $\phi^{\prime}(s x_{l+1,j})$ is unpredictable at training time, and it is not trivial to identify $s$ that could lead to the proper scaling-up effect, as this procedure must be done for each Dropout layer in the network. Additionally, improper scaling can push values outside the effective range of the splines, potentially causing training instabilities \cite{altarabichi2024rethinking,li2024kolmogorov}.

\section{Methods}
%A KAN layer with $n_{in}$ input dimensions and $n_{out}$ output dimensions can be expressed as a 1D vector of functions:
%\begin{equation}
%\Phi = \{\phi_{j,i}\}, \quad i = 1, 2, \ldots, n_{in}, \quad j = 1, 2, \ldots, n_{out}
%\end{equation}
%where $\phi_{j,i}$ are arbitrary functions represented as splines with trainable parameters. We define $x_{i}$ as the pre-activation (input) of the function $\phi_{j,i}$; the post-activation of $\phi_{j,i}$ is denoted by $\tilde{x}_{j,i} = \phi_{j,i}(x_{i})$. The neuron in KAN layer perform a sum of all incoming post-activations to output $y_{j}$.
%\begin{equation}
%y_{j} = \sum_{i=1}^{n_{in}} \tilde{x}_{j,i} =  \sum_{i=1}^{n_{in}} %\phi_{j,i}(x_{i}), \quad j = 1, 2, \ldots, n_{out}
%\label{kan}
%\end{equation}
We design DropKAN to address the previous issues of Dropout by applying a binary dropout mask \(M_{i}\) to the outputs of the activations in training time as observed in Figure~\ref{fig1}. This is different to Dropout where the mask is applied to the output of the neurons $x_{l+1,j}$. We sample a mask to drop some post-activations independently from a Bernoulli distribution with a probability \(p\) of being 1 (indicating that the post-activation is retained) and \(1-p\) of being 0 (indicating that the post-activation is dropped). We will refer to this mode of masking as DropKAN$^{pa}$ (DropKAN post-activation). The output of the KAN layer when DropKAN$^{pa}$ is applied becomes:
\begin{equation}
\begin{aligned}
x^{\prime}_{l+1,j} = \frac{1}{1-p} \sum_{i=1}^{n_{in}} M_{i} \tilde{x}_{j,i} =  \frac{1}{1-p} \sum_{i=1}^{n_{in}} M_{i} \phi_{j,i}(x_{i}), 
j = 1, 2, \ldots, n_{out}
\label{dropkan}
\end{aligned}
\end{equation}
In DropKAN$^{pa}$ the output is scaled-up by a factor of $\frac{1}{1-p}$ to compensate for the dropped post-activations. In test time DropKAN$^{pa}$ behaves exactly as a regular KAN layer as in Equation~\ref{kan}, effectively serving as an identity function. Applying the mask on the post-activations in DropKAN$^{pa}$ combined with scaling-up helps maintain the expected values of the sums in KAN nodes during training, making them approximately the same with and without DropKAN$^{pa}$:
\begin{equation}
\mathbb{E}[\frac{1}{1-p} \sum_{i=1}^{n_{in}} M_{i} \tilde{x}_{j,i}] \approx \mathbb{E} [\sum_{i=1}^{n_{in}} \tilde{x}_{j,i}]
\label{dropkaneq}
\end{equation}
%Unlike with Dropout, DropKAN doesn't mask the nodes after the sum, instead some of the post-activations are masked while the kept ones are scaled-up, which ensure the expected value of the summation taking place in the node with the presence of DropKAN is about the same without DropKAN as observed in Equation~\ref{dropkaneq}.  

It must be noted that the activation function $\phi_{j,i}$ is implemented in \cite{liu2024kan} as the sum of a base function $b(x)$ in addition to the spline function:
\begin{equation}
\phi_{j,i}(x_{i}) = w_{b}b(x_{i}) + w_{s}spline(x_{i})
\label{kanactivation}
\end{equation}
where $w_{b}$ and $w_{s}$ are trainable parameters that control the magnitude of the activation function, and $b(x)=silu(x)={x}/{(1 + e^{-x})}$. The activation function defined in Equation~\ref{kanactivation} allows us to consider an alternative mode of masking the outputs, where the mask $M$ is applied only to the spline functions. We will refer to this mode as DropKAN$^{ps}$ (DropKAN post-spline). Consequently, in the DropKAN$^{ps}$ mode Equation~\ref{kanactivation} becomes:
\begin{equation}
\phi_{j,i}(x_{i}) = w_{b}b(x_{i}) + M_{i}w_{s}spline(x_{i})
\label{dropkanpostspline}
\end{equation}
As suggested by \cite{liu2024kan}, the average value of the spline function is zero because the B-spline coefficients are initialized from $\sim \mathcal{N}(0,\,\sigma^{2})$ with a small variance $\sigma$, leading to $spline(x) \approx 0$. Therefore, the expected value of $\phi_{j,i}(x_{i})$ remains unchanged regardless of whether or not we apply scaling. Therefore, with DropKAN$^{ps}$, we have the flexibility to either scale or not scale the activations. This flexibility is reflected in the following equivalence:
%We propose not scale-up the output of the activations with DropKAN$^{ps}$, unlike what we did with DropKAN$^{pa}$ since:
\begin{equation}    
\begin{aligned}
\mathbb{E}[w_{b}b(x_{i}) + w_{s}spline(x_{i})] \approx 
\mathbb{E} [w_{b}b(x_{i}) + M_{i}w_{s}spline(x_{i})] \approx \mathbb{E} [w_{b}b(x_{i}) + \frac{1}{1-p}M_{i}w_{s}spline(x_{i})]
\label{dropkanpostsplineeq}
\end{aligned}
\end{equation}

%This approximation holds true because, as suggested by \cite{liu2024kan}, the B-spline coefficients are initialized from $\sim \mathcal{N}(0,\,\sigma^{2})$ with a small variance $\sigma$, leading to $spline(x) \approx 0$. Thus, the expected value remains consistent regardless of scaling.

%Equation~\ref{dropkanpostsplineeq} is true since $spline(x) \approx 0$ as \cite{liu2024kan} suggested to initialize the B-spline coefficients from $\sim \mathcal{N}(0,\,\sigma^{2})$ with a small $\sigma$.

%We design DropKAN by randomly masking the output of some of the activations in a KAN layer. The mask in DropKAN is applied prior to the sum operation performed in nodes. DropKAN opertaes on a finer granularity on edges, unlike Dropot that is applied on nodes. Another difference between DropKAN and Dropout is the  

%Consequently DropKAN shares the same intuition of dropout 

\section{Results and Discussion}
This section describes the experimental design we have used to evaluate DropKAN, along with the results obtained in our experiments. The first experiment aimed to evaluate the expected value of a KAN function using DropKAN layers in training mode, and compare to a network with Dropout enabled between the KAN layers. In the second experiment, we compare the performance of a network using DropKAN layers against a regular KANs and KAN regularized with Dropout using a number of classification problems.

\subsection{Experimental Setup}
\label{Experimental Setup}
Our experiments involve 10 popular \cite{altarabichi2021surrogate,altarabichi2023fast} datasets from the UCI Machine Learning Database Repository\footnote[2]{http://archive.ics.uci.edu/ml}. We have included datasets of varying sizes from different domains. Table~\ref{tab1} provides a summary of the number of instances, features and classes of all data sets used in our experiments.

\begin{table*}[t]
\centering
\caption{UCI Data sets used for evaluation.}\label{tab1}
\begin{tabular}{|l|l|l|l|}
\hline
Data set &  No. of Instances & No. of Features &  No. of Classes\\
\hline
dermatology & 366 & 34 & 6\\
german & 1\,000 & 24 & 2\\
semeion & 1\,592 & 265 & 2\\
car & 1\,728 & 6 & 4\\
abalone & 4\,177 & 8 & 28\\
adult & 32\,561 & 14 & 2\\
bank-full & 45\,211 & 16 & 2\\
connect-4 & 67\,556 & 42 & 3\\
diabetes & 101\,766 & 49 & 3\\
census-income & 199\,523 & 41 & 2\\
\hline
\end{tabular}
\label{tab1}
\end{table*}

Each data set is divided into training (60\%), validation (20\%), and testing (20\%) splits. A unified approach of prepossessing is adopted for all data sets, including categorical features encoding, imputation of missing values, and shuffling of instances. Accuracy of the model is the metric we used for the evaluation in all experiments. All reported accuracies are the ones realized on the testing set. 

As for the KANs hyperparameters we have adopted the default values as recommended by \cite{liu2024kan}. The KAN architecture [n$_{in}$, 10, 1] is used across all experiments (unless explicitly mentioned otherwise) for all datasets, where n$_{in}$ is the number of input features in the dataset. The networks  are trained for 2000 steps using Adam optimizer with a learning rate of 0.01, and a batch size of 32. 

In describing the experiments, we will use KANs with five
 different settings:

\begin{itemize}
\item \textbf{No-Drop:} A standard KAN network without any form of masking.
\item \textbf{Dropout$_{w/ scale}$:} A KAN network regularized using a standard dropout layer applied between its layers, with the outputs of the retained neurons scaled up as recommended by \cite{srivastava2014dropout}.
\item \textbf{Dropout$_{w/o\ scale}$:} A standard KAN network regularized using a dropout layer between its layers, without any scaling of the retained neurons.
\item \textbf{DropKAN$^{pa}$:} A KAN network employing DropKAN layers, where the mode is set to mask post activations.
\item \textbf{DropKAN$^{ps}$:} A KAN network employing DropKAN layers, where the mode is set to mask post splines.
\end{itemize}

\subsection{Experiment I - Forward Pass Evaluation}
In this experiment we evaluate the impact of using DropKAN and Dropout on the forward pass of backpropagation for a KAN network. We train the KAN network [6, 2, 2, 1] five times for a total of 100 steps. We carry five forward passes using the validation split on every 10 steps using the five settings of: No-Drop, Dropout$_{w/ scale}$, Dropout$_{w/o\ scale}$, DropKAN$^{pa}_{w/ scale}$, and DropKAN$^{pa}_{w/o\ scale}$. We fix the rate of drop to 0.5 for all drop modes on each layer. We average the value at the output neuron ($x_{3,1}$) to estimate the expected value of the output using the validation split. 

\begin{figure*}[t]
\centering
\subfloat{\includegraphics[width=0.49\textwidth]{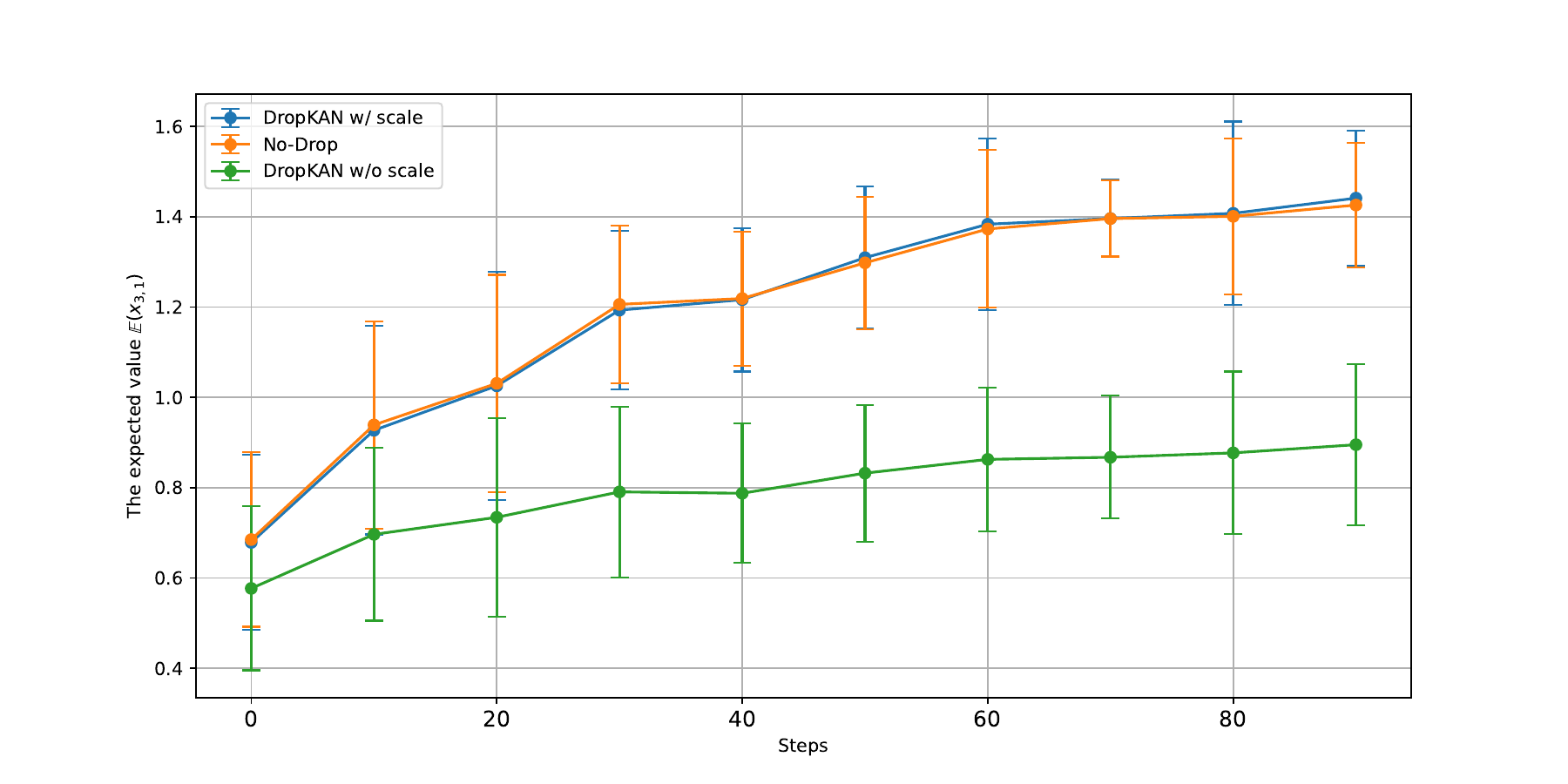}}
\subfloat{\includegraphics[width=0.49\textwidth]{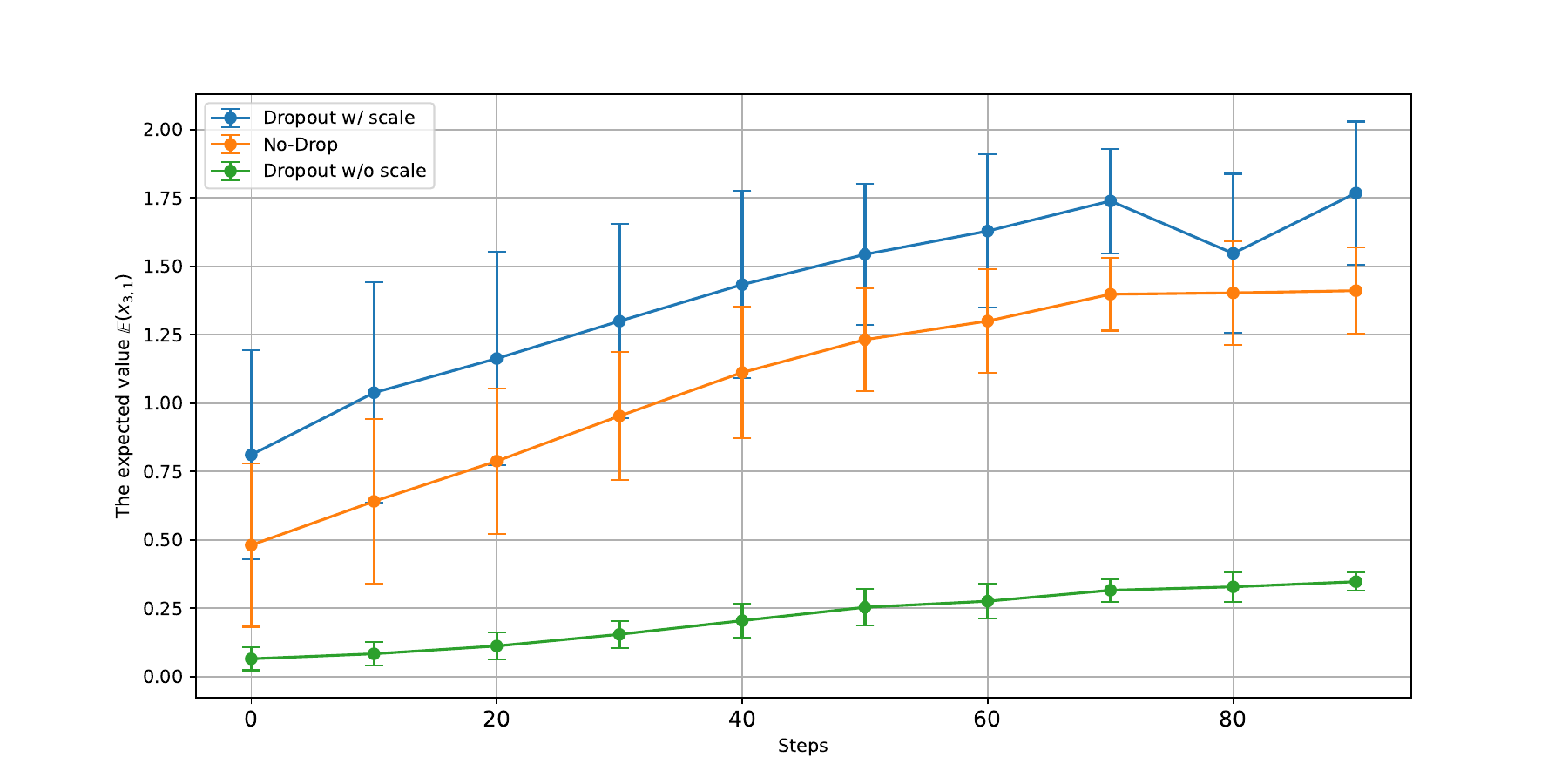}}
\caption{The expected value of the output neuron ($x_{3,1}$) of the [6, 2, 2 ,1] KAN with No-Drop using the {\tt car} dataset compared to DropKAN$^{pa}_{w/ scale}$ and DropKAN$^{pa}_{w/o\ scale}$ (on the left), and to Dropout$_{w/ scale}$ and Dropout$_{w/o\ scale}$ (right) at different stages of the training.}
\label{fig2}
\end{figure*}

The plots in Figure~\ref{fig2} show the results of this experiment, and while it is clear that the propagated signals for both Dropout$_{w/o\ scale}$ and DropKAN$^{pa}_{w/o\ scale}$ become weaker due to the process of zeroing-out, the scaling is allowing DropKAN$^{pa}_{w/ scale}$ to pretty accurately approximate the strength of the signal comparing to the No-Drop mode consistent with Equation~\ref{dropkaneq}. On the other hand, the procedure of scaling-up the input for Dropout is consistently off, and in this case is causing to over-scale the signal. As we have explained earlier, scaling up the input of the non-linear activation function $\phi^{\prime}$ would cause an unpredictable behaviour at training time since $\phi^{\prime}(sx) \neq s\phi^{\prime}(x)$.

\subsection{Experiment II - Classification Problems}
In this experiment, we compare KANs equipped with DropKAN layers against KANs with standard KAN layers and KANs regularized using Dropout. The goal of the experiment is to validate the regularizing effect of DropKAN and to compare it against using standard Dropout with KANs. 

We use a random search to optimize the rates of drop for the DropKAN and Dropout settings, we ran the search for 50 evaluation per setting, and choose the setting with the highest accuracy performance on the validation split. For evaluation we train every setting five times and report the average test accuracy of the runs. 

The findings in Table~\ref{tab2} demonstrate that regularization methods enhanced the performance of the standard KAN (No-Drop) in nine out of ten scenarios, excluding the {\tt car} dataset. Notably, DropKAN variants achieved the highest test accuracy across four datasets for each variant. In contrast, the standard Dropout$_{w/ scale}$ was never the top performer in any dataset. These results indicate that integrating the drop mask within the KAN layer is more effective than applying dropout between layers.

An intriguing observation pertains to the performance of Dropout. The unscaled variant, Dropout$_{w/o\ scale}$, outperformed the standard Dropout in eight out of ten datasets. This supports our earlier assertion that scaling retained neurons can lead to unpredictable performance, as $\phi^{\prime}(sx) \neq s\phi^{\prime}(x)$. Therefore, we recommend avoiding scaling when using Dropout with KANs.

\begin{table*}[t]
\caption{Results of KAN networks trained with Dropout, DropKAN, and no Drop, each result is the mean of five independent runs.}
\begin{center}
\begin{tabular}{|c|c|c|c|c|c|}
\hline
\textbf{Dataset}&\multicolumn{5}{|c|}{\textbf{KANs Accuracy}}\\
\cline{2-6} 
\textbf{} & \textbf{No-Drop} & \textbf{Dropout$_{w/o\ scale}$} & \textbf{Dropout$_{w/ scale}$} & \textbf{DropKAN$^{ps}$} & \textbf{DropKAN$^{pa}$}\\
\hline
dermatology & 73.78\%$ \pm5.94$   & 89.46\%$ \pm5.60$ & 86.49\%$ \pm1.66$ & 92.43\%$ \pm1.21$  & \textbf{92.70\%$ \pm1.54$}\\
\hline
german & 66.70\%$ \pm1.35$& 74.30\%$ \pm2.02$ & 73.40\%$ \pm3.49$ & 70.70\%$ \pm3.03$ &  \textbf{76.6\%$ \pm1.98$}\\
\hline
semeion & 89.66\%$ \pm0.0$ & 94.55\%$ \pm4.50$ & 97.43\%$ \pm0.52$ & \textbf{99.81\%$ \pm0.42$}  & 97.62\%$ \pm0.42$\\
\hline
car & \textbf{91.56\%$ \pm1.18$} & 86.07\%$ \pm1.60$ & 77.69\%$ \pm6.96$ & 89.65\%$ \pm0.63$  & 85.66\%$ \pm2.95$\\
\hline
abalone & 24.98\%$ \pm0.95$ & 23.71\%$ \pm1.96$ & 23.35\%$ \pm1.85$ & 27.68\%$ \pm0.87$  & \textbf{27.85\%$ \pm0.49$}\\
\hline
adult & 84.94\%$ \pm0.17$ & 84.88\%$ \pm0.15$ & 84.66\%$ \pm0.53$ & \textbf{85.27\%$ \pm0.12$} & 85.08\%$ \pm0.19$\\
\hline
bank-full & 90.13\%$ \pm0.37$ & 90.06\%$ \pm0.26$ & 90.12\%$ \pm0.20$ &90.28\%$ \pm0.20$ & \textbf{90.33\%$ \pm0.13$} \\
\hline
connect-4 & 67.26\%$ \pm4.31$ &  70.53\%$ \pm0.39$ &63.85\%$ \pm5.77$ & \textbf{72.14\%$ \pm0.31$} & 68.67\%$ \pm1.72$ \\
\hline
diabetic & 52.56\%$ \pm4.44$ & 56.06\%$ \pm1.58$ & 51.52\%$ \pm2.80$ & \textbf{57.95\%$ \pm0.27$} & 56.55\%$ \pm0.76$\\
\hline
census-income & 94.78\%$ \pm0.21$ &  \textbf{94.90\%$ \pm0.08$} & 94.11\%$ \pm0.23$ & 94.58\%$ \pm0.11$ & 94.69\%$ \pm0.03$ \\
\hline
\end{tabular}
\label{tab2}
\end{center}
\end{table*}

\section{Related Work}
Expanding upon the Dropout technique, various methodologies \cite{wan2013regularization,goodfellow2013maxout,ghiasi2018dropblock} have been proposed to refine the training regularization of Deep Neural Networks (DNNs) for supervised learning. The fundamental idea is to introduce noise into intermediate layers during training. Recent advancements have focused on enhancing regularization in Convolutional Neural Networks (CNNs) through the introduction of structured noise. For instance, the SpatialDropout method \cite{tompson2015efficient} selectively removes entire channels from activation maps, the DropPath scheme \cite{zoph2018learning} opts to discard entire layers, and the DropBlock algorithm \cite{ghiasi2018dropblock} zeros out multiple contiguous regions within activation maps. 

Techniques inspired by Dropout are not limited to ones during the feedforward stage of backpropagation, as in Gradient Dropout proposed in the context of meta-learning by \cite{tseng2020regularizing}, and later generalized by \cite{altarabichi2024rolling} to the supervised training setting, \cite{altarabichi2024rolling} showed that masking the gradient during the backward pass prevents the network from memorizing the data and learning overly simple patterns early in the training, similar ideas could potentially be extended to KANs.

\section{Future Work and Limitations}
It would be interesting to implement and test DropKAN for the newly-emerging KAN-based architectures such as graph  \cite{bresson2024kagnns,kiamari2024gkan,de2024kolmogorov,xu2024fourierkan}, convolutional \cite{bodner2024convolutional}, and transformer \cite{genet2024temporal}. For instance in FourierKAN-GCF (Graph Collaborative Filtering ) \cite{xu2024fourierkan}, regularization strategies like node and message dropout were tested. We believe DropKAN could be seamlessly extended to these architectures. Furthermore, DropKAN could be evaluated with alternative activation function representations beyond B-splines, a prominent research focus. These alternatives include wavelets \cite{bozorgasl2024wav,seydi2024unveiling}, radial basis functions \cite{li2024kolmogorov,ta2024bsrbf}, fractional functions \cite{aghaei2024fkan}, rational functions \cite{aghaei2024rkan}, and sinusoidal functions \cite{reinhardt2024sinekan}.

This paper focused exclusively on evaluating DropKAN within classification scenarios for tabular data. However, there is significant potential for extending DropKAN to other domains where KANs have shown strong performance. The original KAN paper, as presented by Liu et al. \cite{liu2024kan}, highlighted the model's superior performance in data fitting and PDE tasks. Subsequent research has demonstrated the effectiveness of KANs in a variety of areas, including computer vision \cite{cheon2024demonstrating, azam2024suitability}, time series analysis \cite{vaca2024kolmogorov}, engineering design \cite{abueidda2024deepokan}, human activity recognition \cite{liu2024ikan, liu2024initial}, DNA sequence prediction \cite{he2024kan}, and quantum architecture search \cite{kundu2024kanqas}. Future work should explore the application of DropKAN in these diverse fields to fully leverage its potential benefits.

\section{Conclusions}
We present DropKAN an effective method to regularize KANs leading to reliable generalization improvement over the standard KANs, and KANs regularized with Dropout. Our analysis of adapting Dropout to KANs indicates that using it between KAN layers is problematic due to the speciality of the computation graph of KANs. We design DropKAN by
embedding a dropout mask within the KAN layer instead of masking the neurons/nodes. We show that a KAN network constructed using DropKAN layers consistently outperforms a KAN of the same architecture using standard KAN layers, even when the later is regularized with Dropout between its layers.

\bibliography{references}  %%% Remove comment to use the external .bib file (using bibtex).
%%% and comment out the ``thebibliography'' section.

\end{document}